\documentclass{article} 
\usepackage{iclr2026_conference}
\usepackage{times}

\usepackage{hyperref}
\usepackage{url}

\usepackage{titletoc}
\usepackage{tocloft}

\usepackage{amsmath,amssymb}
\usepackage[ruled,vlined,linesnumbered,noend]{algorithm2e}
\usepackage{xcolor}

\usepackage{adjustbox}

\usepackage{wrapfig}
\usepackage{graphicx}
\usepackage{subcaption}
\usepackage{enumitem}

\newcommand{\autocontext}{\textsc{AutoContext}}

\usepackage[most]{tcolorbox} 
\newtcblisting{logbox}[1]{%
  enhanced,
  breakable,                  
  title={#1},                 
  fonttitle=\bfseries,
  coltitle=black,
  colbacktitle=yellow!30,        
  colback=yellow!4,           
  colframe=yellow!50!black,   
  listing only,               
  listing options={           
      basicstyle=\ttfamily\small,
      breaklines=true,
      escapeinside={(*@}{@*)} 
  },
  boxrule=0.5pt,
  arc=0mm,
  left=0.2em,right=0.2em,top=0.2em,bottom=0.2em
}
\usepackage{soul}
\usepackage{booktabs}
\usepackage{multirow}

\newcommand{\hlblue}[1]{\sethlcolor{cyan!20}\hl{#1}}
\newcommand{\hlgreen}[1]{\sethlcolor{green!20}\hl{#1}}

\title{AutoContext: Instance-Level Context Learning for LLM Agents}

\usepackage{authblk}

\author[1]{\textbf{Kuntai Cai}}
\author[2]{\textbf{Juncheng Liu}}
\author[2]{\textbf{Xianglin Yang}}
\author[1]{\textbf{Zhaojie Niu}}
\author[2]{\textbf{Xiaokui Xiao}}
\author[1]{\textbf{Xing Chen}}

\affil[ ]{$^{1}$ByteDance, $^{2}$National University of Singapore}
\affil[ ]{\texttt{\{kuntai.cai, zhaojie.niu, chenxing.xc\}@bytedance.com,
\{xianglin, xkxiao\}@nus.edu.sg, juncheng.liu@u.nus.edu}}

\iclrfinalcopy

\makeatother

\AddToShipoutPicture{}

\makeatletter
\renewcommand{\makevruler}[1][]{} 
\renewcommand{\iclrruler}[1]{}    
\makeatother

\begin{document}
\maketitle

\begin{abstract}
Current LLM agents typically lack \emph{instance-level context}, which comprises concrete facts such as environment structure, system configurations, and local mechanics.
Consequently, existing methods are forced to intertwine exploration with task execution.
This coupling leads to redundant interactions and fragile decision-making, as agents must repeatedly rediscover the same information for every new task.
To address this, we introduce \autocontext{}, a method that decouples exploration from task solving.
\autocontext{} performs a systematic, one-off exploration to construct a reusable knowledge graph for each environment instance.
This structured context allows off-the-shelf agents to access necessary facts directly, eliminating redundant exploration.
Experiments across \textsc{TextWorld}, \textsc{ALFWorld}, \textsc{Crafter}, and \textsc{InterCode-Bash} demonstrate substantial gains: for example, the success rate of a ReAct agent on \textsc{TextWorld} improves from 37\% to 95\%, highlighting the critical role of structured instance context in efficient agentic systems.
\end{abstract}

\section{Introduction}

Large language model (LLM) agents are increasingly deployed in interactive environments such as robotic control~\citep{brohan2023can,guo2025seed1}, game playing~\citep{wangvoyager, zhu2023ghost, luintelligent}, web navigation~\citep{chae2025web,yao2022webshop}, and operating systems~\citep{xie2024osworld}, where the agent must act, observe, and adapt over long horizons.
In such settings, many facts that determine task success depend on the specific environment instance, such as spatial layouts, file system hierarchies, and the precise behaviors of tools.
To gather such information, humans typically perform an environment familiarization process before attempting a task, which helps them build a mental model of the environment instance and facilitates efficient task execution.
LLM agents, in contrast, rarely receive such instance-specific context, leading to degraded performance and inefficiency.
This paper introduces \autocontext{}, a novel method that systematically explores an environment instance and produces reusable instance context (a knowledge graph), which can be plugged into downstream agents to significantly improve their performance and efficiency.

\begin{figure*}[t]
  \centering
  \vspace{-3mm}
  \includegraphics[width=1.0\linewidth]{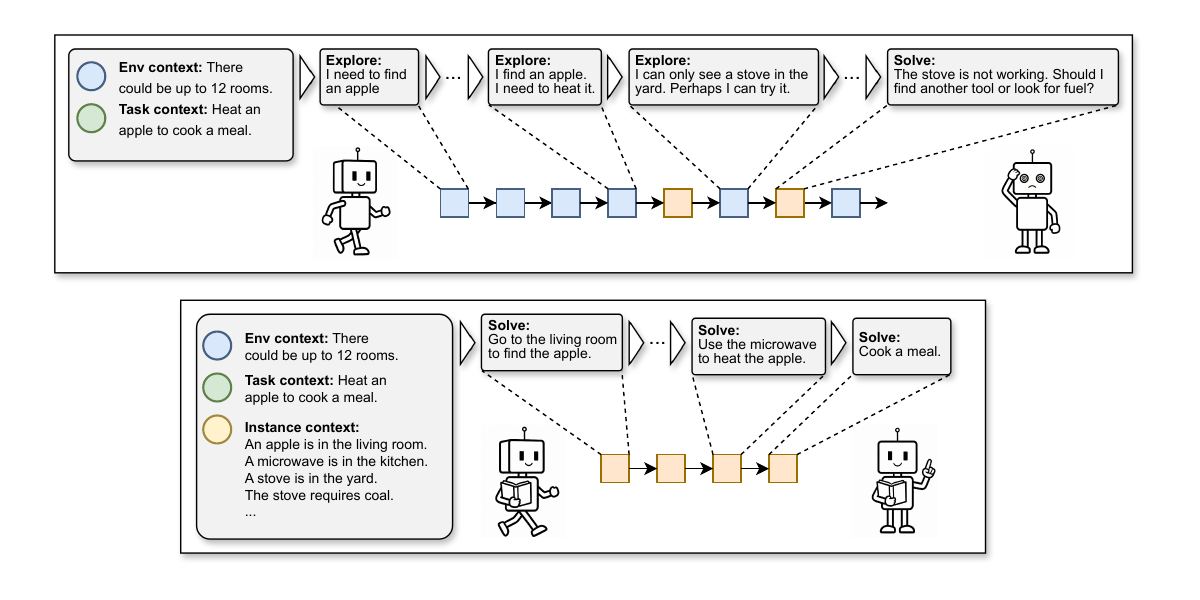}
  \vspace{-3mm}
  \caption{\autocontext{} replaces repeated, agent-specific exploration with a reusable instance-level context. \textbf{Top:} Existing methods often rely on intertwined explore-and-solve steps, which demand substantial effort to design tightly coupled, ad hoc exploration strategies for each agent and still force agents to repeatedly rediscover instance-specific facts. \textbf{Bottom:} \autocontext{} performs a single, one-off exploration to construct a shared instance context that can be reused across tasks and agents, allowing downstream agents to bypass redundant exploration and focus on solving tasks directly.}
  \label{fig:demo-diff}
\end{figure*}

Existing methods typically guide agents using two forms of auxiliary information (see Figure~\ref{fig:demo-diff}):
(1) \emph{Environment-level context}, which captures global mechanisms shared across the domain~\citep{qiao2024agent, chae2025web, fu2024autoguide}; and
(2) \emph{Task-level context}, which provides goal-specific instructions or demonstrations~\citep{prasad2024adapt, zhaoempowering}.
However, these methods lack explicit \emph{instance-level context}, i.e., the specific, local facts that define the current environment instance.

To compensate for this missing piece, existing works are forced into an \emph{explore-and-solve paradigm}, relying on extensive online interaction to uncover instance-specific details during task execution.
Specifically, some approaches employ LLM-guided tree searches to progressively discover and return to promising states~\citep{luintelligent, golchha2024language, yao2023tree}, while others rely on iterative trial-and-error procedures to make progress through repeated attempts~\citep{chen2024automanual, zhao2024expel, AnokhinSSEK0B25, shinn2023reflexion}.
Despite their effectiveness, these intertwined explore-and-solve approaches suffer from three limitations.
First, their exploration strategies are often ad hoc and tightly coupled to specific agent architectures, imposing substantial design and implementation costs.
Second, the knowledge accumulated during exploration (e.g., saved search trees or trajectories) is method-specific, making it difficult to reuse across different agents.
Third, the exploration effort is ephemeral: since the gathered information is rarely persisted in a durable format, agents must conduct redundant exploration even when returning to the same environment for a new task.

To address these limitations, we introduce \emph{Instance-Level Context Learning (ILCL)}, which aims to perform a compact, one-off exploration that extracts reusable, task-agnostic facts into a durable document $D_e$.
Downstream agents can then use $D_e$ to avoid redundant exploration and operate with a clearer view of the environment's latent structure (Figure~\ref{fig:demo-diff}, bottom). We propose \autocontext{}, a simple yet effective method for ILCL. It uses a \emph{TODO forest} to manage exploration states and a \emph{Plan–Act–Extract} loop to construct a portable knowledge graph. Unlike solver-coupled approaches, \autocontext{} produces a plug-in context layer that requires no customizations and can be seamlessly serialized and appended to the prompts of diverse downstream agents.

We evaluate \autocontext{} on four benchmarks: \textsc{TextWorld}~\citep{cote2018textworld}, \textsc{ALFWorld}~\citep{shridharalfworld}, \textsc{Crafter}~\citep{hafnerbenchmarking}, and \textsc{InterCode-Bash}~\citep{yang2023intercode}.
By equipping existing state-of-the-art agents with the resulting context $D_e$, we observe remarkable improvements in both success rates and interaction efficiency across all domains.
For example, on \textsc{TextWorld}, the success rate of ReAct~\citep{yao2023react} improves from 37\% to 95\%, and IGE~\citep{luintelligent} improves from 81\% to 95\%.
These results demonstrate that \autocontext{} empowers agents with the necessary instance-specific context, and that a lightweight, pre-task exploration is often sufficient to yield substantial gains across diverse environments.

\section{Related Work}
\label{sec:related-work}

\paragraph{Task-level Knowledge Learning.}
A substantial line of research~\citep{wangvoyager, zhangomni, guan2024amor, zhu2025knowagent, qiao2024agent, chen2024automanual, zhao2024expel, fu2024autoguide, basavatia2024starling, kirk2024improving, shinn2023reflexion, chenscaling, AgentRM, wu2023spring} investigates how agents acquire task-specific rules. 
For instance, AutoManual~\citep{chen2024automanual} induces rules via trial-and-error, and ExpeL~\citep{zhao2024expel} distills experiences into insights, both aiming to optimize the agent's policy or domain-level mechanics. In contrast, \autocontext{} captures instance-level facts, providing a reusable knowledge base that complements these strategy-oriented methods.

\paragraph{LLM-based Exploration.}
Another line of work~\citep{luintelligent, golchha2024language, song2024trial, synworld, du2023guiding} studies how LLM agents can be used to improve exploration. 
Intelligent Go-Explore (IGE)~\citep{luintelligent} leverages LLM’s internalized knowledge to archive promising states and resume exploration from them. 
Language Guided Exploration (LGE)~\citep{golchha2024language} uses LLMs to propose promising next actions. 

These methods enhance task performance by leveraging improved exploration strategies to perform an exhaustive search, but they do not yield lasting knowledge about the environment instance. We adopt a similar assumption as their setting, that the agent can return to a previously saved state, achieved either via a simulator's reset function or by prompting the agent with a past trajectory.

\paragraph{Instance Memory.}
Several approaches~\citep{gao2025efficient, holt2025improving, kagaya2024rap, huang2024wese, ammanabrolugraph, AnokhinSSEK0B25, hu2025hiagent} augment agents with external memory or knowledge bases to store information gathered during task execution.
For example, HiAgent~\citep{hu2025hiagent} saves working memory by subgoals, and retrieves detailed trajectories only when necessary to control the context window. CDMem~\citep{gao2025efficient} proposes a framework that organizes trajectories into short- and long-term memories and retrieves relevant items to support task completion. While these approaches retain instance-dependent facts, their exploration remains task-driven. Consequently, the acquired knowledge tends to be fragmented and biased toward the specific trajectories taken. In contrast, our work constructs a comprehensive and reusable instance-level context that is decoupled from any specific task.

Additional related work is provided in Appendix~\ref{sec:more-related-work}.

\section{Problem Setting}\label{sec:problem}

We introduce the problem of \emph{Instance-Level Context Learning (ILCL)}. Formally, let $\mathcal{E}$ denote a class of environments. Each instance $e \in \mathcal{E}$ can be modeled as a Partially Observable Markov Decision Process (POMDP): $e=\langle \mathcal{S}, \mathcal{A}, T, \mathcal{O}\rangle$, where $\mathcal{S}$ is the state space, $\mathcal{A}$ is the action space, $T$ is the transition function, and $\mathcal{O}$ is the observation space. The instance $e$ is governed by persistent but initially hidden facts, such as environment structure, system configurations, and local mechanics, which can be uncovered only after reaching certain states.

User tasks $t$ are drawn from a distribution $P_e$ defined over the instance $e$, where each task starts from the same initial state $s_0 \in \mathcal{S}$, and is solved by an agent policy $\pi_t$.
We write $\pi_{t \mid D_e}$ to denote the solver $\pi_t$ operating with access to a document $D_e$.
Let $U_t(\cdot)$ denote the utility function of task $t$. Our objective is to perform a one-off exploration on $e$ to construct a reusable context $D_e$ that maximizes the amortized utility of all tasks:
\begin{equation}
  \label{eq:obj}
  J(D_e)
  = \mathbb{E}_{t \sim P_e} \left[ U_t\big(\pi_{t \mid D_e}\big) \right].
\end{equation}
Specifically, we use success rates and interaction steps to measure the utility $U_t$ in our experiments (Section~\ref{sec:exp}).

While Equation~\ref{eq:obj} formally captures our objective, directly optimizing it is intractable as the task distribution $P_e$ is typically unknown during exploration. Consequently, we adopt a task-agnostic surrogate: maximizing the coverage of the environment's latent structure. Our objective is to exhaustively discover hidden instance-specific facts that serve as prerequisites for potential tasks. To achieve this, we materialize $D_e$ as a structured knowledge graph $G_e$. This format captures atomic facts in a portable manner, allowing them to be continuously updated during exploration and serialized into prompts for LLM processing.

\begin{figure*}[t]
    \centering
    \includegraphics[width=1.0\linewidth]{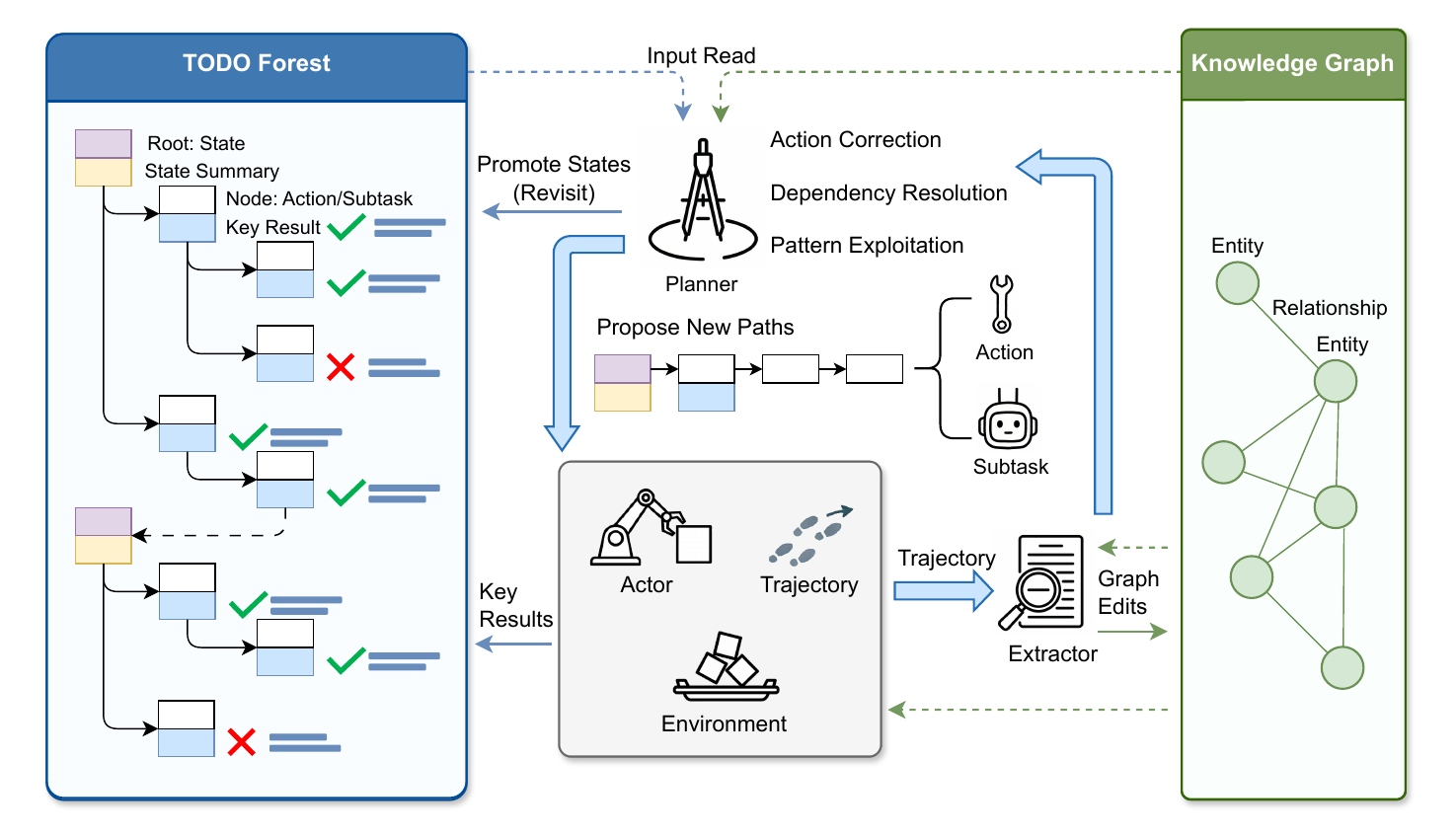}
    \vspace{-2mm}
    \caption{\textbf{Overview of AutoContext.}}
    \vspace{-0mm}
    \label{fig:overview}
\end{figure*}

\section{Method}

We introduce \autocontext{}, a task-agnostic method for ILCL. Given an environment instance, \autocontext{} performs a compact exploration to construct a reusable and comprehensive knowledge graph.
As illustrated in Figure~\ref{fig:overview}, the method relies on two main components.
The \emph{TODO forest} serves as the core data structure, organizing states and subtasks into trees to provide a compact exploration representation.
The \emph{Plan--Act--Extract Loop} iteratively interacts with the environment driven by the forest, populating the knowledge graph until coverage is sufficient or the budget is exhausted.

\subsection{TODO Forest}\label{sec:todo-forest}

In the \emph{TODO Forest}, each tree is rooted at a \emph{state}, which represents a snapshot of the environment and is annotated with a brief corresponding summary.
Within a tree, nodes represent actions or subtasks performed from that state.
This structure creates a global view of the exploration history that can be serialized into indented text for the LLM to process without overwhelming the context window.
The forest operates in two distinct modes depending on the granularity of the environment:

\paragraph{Action Mode.}
In this mode, each non-root node represents an executable atomic action.
Consequently, a path from the root to a specific node constitutes a trajectory.
Each action node is paired with a \emph{key result}, a compact abstraction of the observation.
By explicitly recording both successful outcomes and negative feedback (e.g., unsatisfied preconditions or syntax errors), the forest serves as a history of valid state transitions and infeasible attempts.

\paragraph{Agent Mode.}
This mode addresses complex environments with long horizons where incorporating the full history is prohibitive.
In this mode, non-root nodes correspond to high-level subtasks, formally represented as \texttt{agent("task description")}.
Each node preserves only a concise summary of the final outcome as the key result, shielding \autocontext{} from raw, verbose trajectories.
This hierarchical design enables the model to prioritize high-level task dependencies over low-level control details.

\subsection{Plan-Act-Extract Loop}
\label{sec:plan-act-extract loop}

\autocontext{} builds the instance knowledge graph through a loop driven by the TODO forest. Each iteration executes three LLM-based agents to expand the forest and refine the knowledge graph.
Prompt templates are provided in Appendix~\ref{sec:prompt}.

\paragraph{Planner: Proposing TODOs.}
The Planner is prompted to propose new paths (i.e., action sequences in action mode, or task sequences in agent mode) given the current TODO forest and the knowledge graph.
We employ three prompting strategies targeting paths with different objectives:
\begin{itemize}
    \item \textbf{Action Correction:} The Planner is prompted to scan the TODO forest for failed attempts and examine the negative feedback in the key results. It is required to avoid all failed attempts, diagnose the errors (e.g., incorrect syntax or invalid arguments), and propose reformulated actions to resolve execution failures.
    \item \textbf{Dependency Resolution:} The Planner is prompted to identify failures caused by unsatisfied preconditions. It constructs logical action sequences to establish these requirements (e.g., preparing necessary files or materials). Particularly, in agent mode, the Planner decomposes complex objectives or reorganizes subtasks to ensure a feasible execution order and resolve high-level dependencies.
    \item \textbf{Pattern Exploitation:} The Planner is prompted to infer potentially discoverable information based on the patterns of successful attempts in the current TODO forest. It scans the knowledge graph for missing details and systematically interacts with the environment to fill these gaps.
\end{itemize}
Finally, these proposed paths are validated against the forest to ensure they originate from existing states and avoid redundancy.

\paragraph{Actor: Completing TODOs.} The Actor executes the proposed TODOs and returns the resulting trajectories. For each proposed path, the Actor first restores the environment to the explored part of the path. Then, it executes the new steps: either performing atomic actions directly (action mode) or instantiating a ReAct agent to iteratively complete the subtask (agent mode). Upon completion, the trajectory and an environment checkpoint are stored at each new node, allowing it to serve as a starting point for future restoration. Finally, the resulting trajectories are processed by LLMs to update the key results in the forest for those TODO paths.

\paragraph{Extractor: Updating the Knowledge Graph.} The Extractor updates the knowledge graph based on the trajectories returned by the Actor. Leveraging the internal knowledge of the LLM, we allow for the construction of a free-form knowledge graph without specifying a fixed schema. Specifically, we prompt the Extractor to extract atomic facts from the trajectories in a triplet format. Based on these facts, it generates a list of edits to add or remove triplets in the knowledge graph. These edits are then applied to the graph to incorporate the newly discovered information.

\paragraph{Planner: Proposing States.}
After updating the knowledge graph, the Planner revisits the forest to promote nodes into new root states.
Nodes that uncover novel information or establish necessary preconditions for deep exploration are prioritized.
For each promoted state, the LLM is prompted to generate a summary derived from the corresponding trajectory.
This strategy maintains the forest structure by keeping individual trees shallow, ensuring the context remains concise and readable for the LLM.

\paragraph{Loop Termination.}
Finally, the exploration loop continues until the Planner determines that the environment has been sufficiently explored, or when the maximum iteration budget is reached.

\section{Experimental Evaluation}
\label{sec:exp}

Our experiments aim to answer the following main research questions.
\begin{itemize}[leftmargin=*]
    \item \textbf{RQ1:} How much performance gain can be achieved by \autocontext{}?
    \item \textbf{RQ2:} How efficient are \autocontext{} and the baselines with the instance-level context?
    \item \textbf{RQ3:} What is the contribution of each component of \autocontext{}?
\end{itemize}
We also provide a case study for \autocontext{} in Appendix~\ref{sec:case study}.

\subsection{Experiment Setup}

\paragraph{Benchmarks.} We evaluate \autocontext{} on four diverse benchmarks: \textsc{InterCode-Bash}~\citep{yang2023intercode}, \textsc{TextWorld}~\citep{cote2018textworld}, \textsc{ALFWorld}~\citep{shridharalfworld}, and \textsc{Crafter}~\citep{hafnerbenchmarking}. \textsc{InterCode-Bash} serves as a testbed for OS-level control, comprising 24 testing tasks where agents operate within a Docker container to answer system queries and manage file systems. Following~\citet{luintelligent}, we adopt the cooking challenge in \textsc{TextWorld}, which consists of 25 randomly generated environments. This benchmark is deliberately challenging, requiring agents to navigate up to 12 rooms to locate ingredients and execute multi-step recipes. For \textsc{ALFWorld}, a benchmark aligning text-based interactions with embodied household tasks, we use the 134 unseen test environments. \textsc{Crafter} provides an open-ended survival world where agents must gather resources to progress through a technology tree and unlock advanced achievements. We use 40 randomly generated environments for evaluation. We apply \autocontext{} in action mode for \textsc{InterCode-Bash}, \textsc{TextWorld}, and \textsc{ALFWorld}, while adopting agent mode for \textsc{Crafter} to better handle its long horizons.

\paragraph{Baselines.}
We incorporate \autocontext{} into a wide range of existing state-of-the-art methods to verify its broad applicability.
By construction, the knowledge graph produced by \autocontext{} is both task-agnostic and agent-agnostic, and is simply serialized and appended to the prompts of baselines without any customization.
The baselines are: (1) Intelligent-Go-Explore (IGE)~\citep{luintelligent}: an LLM-driven go-explore method, which searches the environment by archiving interesting states and continuing exploration from those states. 
(2) ReAct~\citep{yao2023react}: A classic Chain-of-Thought method, which prompts the LLM to generate a reasoning process. 
(3) AutoManual~\citep{chen2024automanual}: A task-level context generation method, with advanced architecture for completing tasks by LLM-generated code. 
(4) Reflexion~\citep{shinn2023reflexion}: A method that generates self-reflections from previous trials to improve future decisions. We limit Reflexion to a maximum of three trials.
(5) RAP~\citep{kagaya2024rap}: A memory framework that dynamically retrieves relevant past successful trajectories to guide planning via in-context learning.
(6) HiAgent~\citep{hu2025hiagent}: An instance memory method that stores collected information as summarized memory chunks to manage working memory, while retaining a retrieval module for accessing detailed trajectories.

\paragraph{LLMs} Unless noted, all methods are driven by DeepSeek-V3 (i.e., \texttt{DeepSeek-V3-0324}~\citep{liu2024deepseek}), except that for \textsc{Crafter} we employ \texttt{DeepSeek-R1-0528} to construct knowledge graphs. We leverage R1's inference-time reasoning capability to better handle the exploration complexity of \textsc{Crafter}. This setup allows us to investigate how lightweight models can be leveraged for downstream evaluation tasks, while more computationally intensive models are reserved for generating high-quality knowledge graphs. We also employ GPT-4.1~\citep{openai2025gpt41} to assess the robustness of our methods across different LLMs.

\subsection{Experimental Results}
\label{sec:experimental-results}

\paragraph{RQ1: How much performance gain can be achieved by \autocontext{}?} We first examine whether augmenting existing baselines with \autocontext{} yields significant improvements across benchmarks. For \textsc{TextWorld}, \textsc{ALFWorld}, and \textsc{InterCode-Bash}, we report the success rates to measure performance. Specifically, for \textsc{TextWorld} and \textsc{ALFWorld}, we evaluate methods under varying environment step limits (including an unlimited setting). For \textsc{Crafter}, we follow the benchmark’s official evaluation protocol, reporting scores computed as the geometric mean of achievement completion rates across all environments.

\begin{table*}[t]
\vspace{-1mm}
\small
\caption{Success rates (\%) on \textsc{TextWorld} under increasing step budgets.}
\vspace{-3mm}
\label{tab:textworld}
\begin{center}
\begin{adjustbox}{max width=\textwidth}
\begin{tabular}{llccccc}
\toprule
Model & Method & \bf 50 & \bf 100 & \bf 400 & \bf 1600 & \bf unlimited \\
\midrule
\multirow{8}{*}{deepseek} & ReAct & 16 $\pm$ 7 & 35 $\pm$ 7 & 37 $\pm$ 5 & 37 $\pm$ 5 & 37 $\pm$ 5 \\
 & ReAct + \autocontext{} & \textbf{78 $\pm$ 6} & 95 $\pm$ 2 & 95 $\pm$ 2 & 95 $\pm$ 2 & 95 $\pm$ 2 \\
 & Reflexion & 16 $\pm$ 11 & 31 $\pm$ 3 & 45 $\pm$ 4 & 45 $\pm$ 4 & 45 $\pm$ 4 \\
 & Reflexion + \autocontext{} & 73 $\pm$ 5 & \textbf{96 $\pm$ 3} & \textbf{99 $\pm$ 2} & \textbf{99 $\pm$ 2} & \textbf{99 $\pm$ 2} \\
 & IGE & 0 $\pm$ 0 & 0 $\pm$ 0 & 36 $\pm$ 6 & 79 $\pm$ 7 & 81 $\pm$ 8 \\
 & IGE + \autocontext{} & 0 $\pm$ 0 & 0 $\pm$ 0 & 72 $\pm$ 3 & 95 $\pm$ 2 & 95 $\pm$ 2 \\
 & HiAgent & 1 $\pm$ 2 & 12 $\pm$ 3 & 47 $\pm$ 8 & 49 $\pm$ 7 & 49 $\pm$ 7 \\
 & HiAgent + \autocontext{} & 8 $\pm$ 3 & 41 $\pm$ 8 & 76 $\pm$ 11 & 77 $\pm$ 10 & 79 $\pm$ 9 \\
\cmidrule{1-7}
\multirow{4}{*}{GPT4.1} & ReAct & 33 $\pm$ 7 & 81 $\pm$ 2 & 91 $\pm$ 8 & 93 $\pm$ 7 & 93 $\pm$ 7 \\
 & ReAct + \autocontext{} & \textbf{83 $\pm$ 5} & \textbf{100 $\pm$ 0} & \textbf{100 $\pm$ 0} & \textbf{100 $\pm$ 0} & \textbf{100 $\pm$ 0} \\
 & Reflexion & 35 $\pm$ 5 & 77 $\pm$ 3 & \textbf{100 $\pm$ 0} & \textbf{100 $\pm$ 0} & \textbf{100 $\pm$ 0} \\
 & Reflexion + \autocontext{} & \textbf{83 $\pm$ 2} & \textbf{100 $\pm$ 0} & \textbf{100 $\pm$ 0} & \textbf{100 $\pm$ 0} & \textbf{100 $\pm$ 0} \\
 
 & IGE & 0 $\pm$ 0 & 0 $\pm$ 0 & 43 $\pm$ 12 & 69 $\pm$ 8 & 72 $\pm$ 7 \\
 & IGE + \autocontext{} & 0 $\pm$ 0 & 0 $\pm$ 0 & 81 $\pm$ 2 & \textbf{100 $\pm$ 0} & \textbf{100 $\pm$ 0} \\
\bottomrule
\end{tabular}
\end{adjustbox}
\vspace{-1mm}
\end{center}
\end{table*}

\begin{table*}[t]
\small
\caption{Success rates (\%) on \textsc{ALFWorld} under increasing step budgets.}
\vspace{-3mm}
\label{tab:alfworld}
\begin{center}
\begin{tabular}{lccccc}
\hline
\bf Method & \bf 5 & \bf 10 & \bf 40 & \bf 160 & \bf unlimited \\
\hline
ReAct & 13.1 $\pm$ 1.1 & 48.3 $\pm$ 3.6 & 76.5 $\pm$ 1.0 & 77.5 $\pm$ 0.8 & 77.5 $\pm$ 0.8 \\
ReAct + \autocontext{} & 26.7 $\pm$ 1.5 & \textbf{97.3 $\pm$ 1.7} & 98.0 $\pm$ 1.1 & 98.0 $\pm$ 1.1 & 98.0 $\pm$ 1.1 \\
IGE & 10.4 $\pm$ 0.7 & 10.4 $\pm$ 0.7 & 56.3 $\pm$ 4.4 & 86.2 $\pm$ 1.9 & 92.6 $\pm$ 1.3 \\
IGE + \autocontext{} & \textbf{27.9 $\pm$ 0.4} & 29.1 $\pm$ 1.1 & 93.3 $\pm$ 1.3 & 97.3 $\pm$ 0.4 & 98.5 $\pm$ 0.7 \\
AutoManual & 3.4 $\pm$ 0.4 & 28.8 $\pm$ 0.7 & 94.8 $\pm$ 0.5 & 97.9 $\pm$ 0.5 & 97.9 $\pm$ 0.5 \\
AutoManual + \autocontext{} & 17.6 $\pm$ 1.1 & 83.2 $\pm$ 1.2 & \textbf{99.7 $\pm$ 0.4} & \textbf{99.7 $\pm$ 0.4} & \textbf{99.7 $\pm$ 0.4} \\
RAP & 5.4 $\pm$ 2.6 & 48.8 $\pm$ 1.2 & 85.4 $\pm$ 2.2 & 86.1 $\pm$ 3.0 & 86.1 $\pm$ 3.0 \\
RAP + AutoContext & 10.1 $\pm$ 0.4 & 81.7 $\pm$ 1.1 & 96.0 $\pm$ 1.5 & 96.0 $\pm$ 1.5 & 96.0 $\pm$ 1.5 \\
\hline
\end{tabular}
\end{center}
\vspace{-2mm}
\end{table*}

\subparagraph{TextWorld.}
Table~\ref{tab:textworld} reports the mean success rates across varying step budgets. 
ReAct faces significant challenges on this benchmark, achieving only 37\% success even with unlimited steps, largely due to frequent navigation errors. 
However, when equipped with the instance context from \autocontext{}, its performance substantially improves: the success rate under a 50-step budget rises from 16\% to 78\%, and further reaches 95\% with an unlimited budget. 
We observe similar gains for IGE. 
Although IGE, as a search-oriented method, typically outperforms ReAct given sufficient steps, it remains constrained by the limited context window. 
It is forced to discard states during exploration, which leads to information loss and redundant revisits. 
\autocontext{} addresses this by maintaining a compact TODO forest of explored trajectories. 
Consequently, IGE improves from 36\% to 72\% under a 400-step budget, and from 81\% to 95\% with unlimited steps. 
These results indicate that the knowledge graph serves not only as an efficiency booster but also as a compact environmental model, enabling agents to handle complex structures effectively within a limited context window.

\subparagraph{ALFWorld.} Consistent with TextWorld, \autocontext{} delivers notable improvements across all baselines (Table~\ref{tab:alfworld}). Specifically, under a tight budget of 10 steps, ReAct's success rate increases from 48.3\% to 97.3\%, and AutoManual rises from 28.8\% to 83.2\%. We also observe substantial gains for RAP (48.8\% to 81.7\%). However, the marginal improvement for RAP is slightly smaller compared to simpler agents like ReAct. This is likely because RAP already incorporates rich context via retrieved trajectories; appending the knowledge graph may lead to information saturation within the prompt, whereas simpler agents can absorb the structured context without competing for the model's attention span. These significant performance gains demonstrate that once instance-level knowledge is consolidated into a structured graph, agents can bypass redundant exploration and focus on executing complex embodied tasks directly with near-optimal efficiency.

\begin{figure}[t]
  \centering
  \begin{subfigure}[t]{0.33\linewidth}
    \centering
    \includegraphics[width=\linewidth]{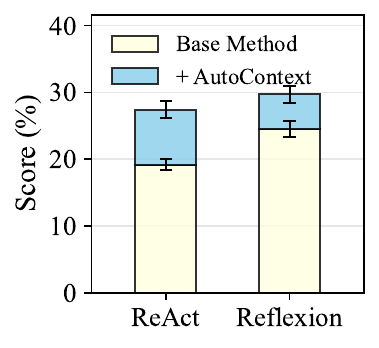}
    \vspace{-5mm}
    \subcaption{\textsc{Crafter}}
    \label{fig:crafter-main-results}
  \end{subfigure}
  \hspace{2mm}
  \begin{subfigure}[t]{0.33\linewidth}
    \centering
    \includegraphics[width=\linewidth]{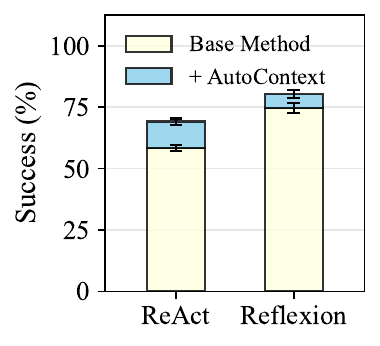}
    \vspace{-5mm}
    \subcaption{\textsc{InterCode-Bash}}
    \label{fig:intercode-main-results}
  \end{subfigure}
  \vspace{-0mm}
  \caption{Results on \textsc{Crafter} and \textsc{InterCode-Bash}}
  \label{fig:crafter-intercode}
  \vspace{0mm}
\end{figure}

\subparagraph{Crafter.}
The official scores, averaged over four runs with 95\% confidence intervals, are presented in Figure~\ref{fig:crafter-main-results}.
\autocontext{} yields marked improvements for both ReAct and Reflexion, demonstrating its generality and effectiveness in open-ended survival environments.
\autocontext{} leverages a TODO forest to systematically explore the technology tree, thereby identifying the precise sequence of prerequisites required for complex actions.
Once these underlying rules and critical locations are consolidated into the context document, ReAct and Reflexion can utilize this structured knowledge to plan and execute long-horizon strategies.
In contrast, without such context, baseline agents struggle to formulate strategic plans, often resorting to aimless exploration.
In an environment where resources are scarce and survival is strictly constrained, this inefficiency inevitably leads to lower scores.

\begin{figure}[t]
  \centering
  \begin{subfigure}[t]{\linewidth}
    \centering
    \vspace{0mm}
    \includegraphics[width=0.65\linewidth]{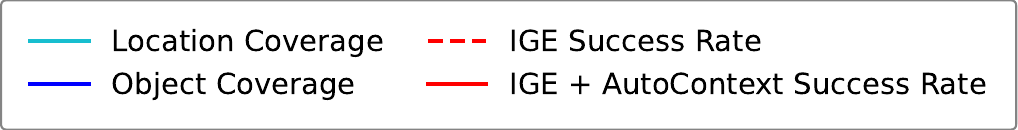}
    \vspace{-0mm}
  \end{subfigure}
  \par\smallskip
  \begin{subfigure}[t]{0.36\linewidth}
    \centering
    \includegraphics[width=\linewidth]{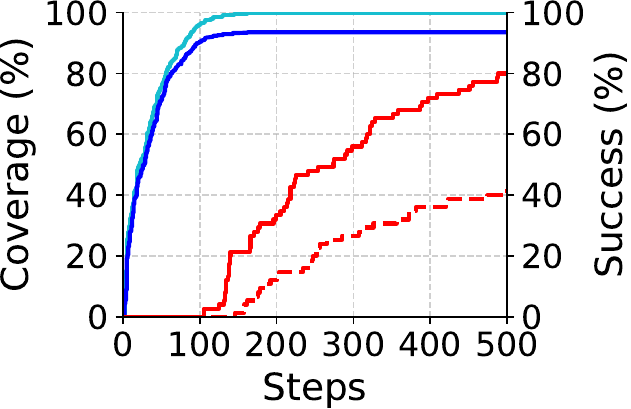}
    \subcaption{\textsc{TextWorld}}
    \label{fig:coverage1}
  \end{subfigure}
  \hspace{3mm}
  \begin{subfigure}[t]{0.36\linewidth}
    \centering
    \includegraphics[width=\linewidth]{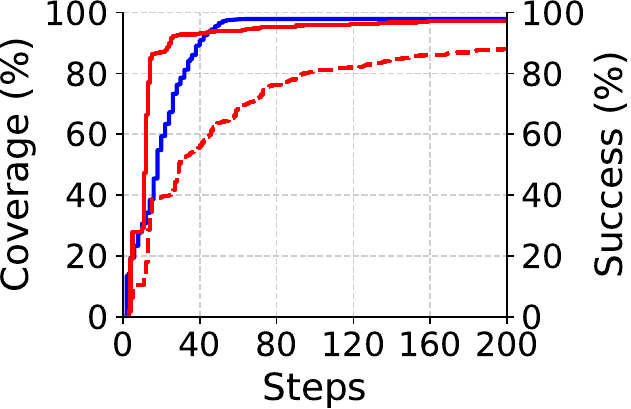}
    \subcaption{\textsc{ALFWorld}}
    \label{fig:coverage2}
  \end{subfigure}
  \vspace{-0mm}
  \caption{Convergence rates and success rates vs. steps}
  \label{fig:coverage}
  \vspace{-0mm}
\end{figure}

\subparagraph{InterCode-Bash.}

The results, averaged over 40 runs with 95\% confidence intervals, are presented in Figure~\ref{fig:intercode-main-results}.
\autocontext{} achieves significant performance gains across \textsc{InterCode-Bash} tasks by synthesizing a comprehensive reference manual that covers both file system hierarchies and tool specifications.
An example is the handling of symbolic links: without instance-level context, ReAct and Reflexion often inadvertently filter out these links when selecting files relevant to a query, simply because they are unaware of their existence.
By explicitly listing the file structure, \autocontext{} enables agents to correctly identify and utilize these links.
Furthermore, we observe that agents relying solely on internal parametric knowledge often generate command arguments that conflict with the specific software versions installed in the environment.
The tool documentation provided by \autocontext{} aligns the agent's generation process with the actual environment, resolving these version mismatches and substantially improving the success rate of command execution.

\paragraph{RQ2: How efficient are \autocontext{} and the baselines with the instance-level context?}

\begin{table*}[t]
  \centering
  \caption{Average steps of successful runs.}
  \label{tab:avg steps}
  \vspace{-0mm}
  
  \begin{subtable}[t]{0.48\textwidth}
  \centering
  \caption{\textsc{TextWorld}.}
  \label{tab:average steps textworld}
  \vspace{-0mm}
  \begin{tabular}{lcc}
  \hline
  \bf Method & \bf Baseline & \bf + \autocontext{} \\
  \hline
  ReAct & 60.7 & \bf 42.7 \\
  Reflexion & 87.2 & \bf 48.5 \\
  IGE & 594.5 & \bf 320.5 \\
  HiAgent & 214.6 & \bf 177.5 \\
  \hline
  \end{tabular}
  \end{subtable}
  \hfill
  \begin{subtable}[t]{0.48\textwidth}
  \centering
  \caption{\textsc{ALFWorld}.}
  \label{tab:average steps alfworld}
  \vspace{-0mm}
  \begin{tabular}{lcc}
  \hline
  \bf Method & \bf Baseline & \bf + \autocontext{} \\
  \hline
  ReAct & 11.4 & \bf 6.6 \\
  IGE & 60.6 & \bf 13.4 \\
  AutoManual & 17.4 & \bf 8.4 \\
  RAP & 11.2 & \bf 8.2 \\
  \hline
  \end{tabular}
  \end{subtable}
  
  \vspace{-0mm}
  \end{table*}
  
  \begin{table*}[t!]
  \centering
  \vspace{-0mm}
  \caption{Ablation study of \autocontext{} components.}
  \vspace{-0mm}
  \label{tab:ablation}
  \begin{tabular}{lcc}
  \hline
  \textbf{Method} & \textbf{TextWorld} & \textbf{ALFWorld} \\
  \hline
  ReAct + \autocontext{} (Ours) & 95 $\pm$ 2 & 98.0 $\pm$ 1.1 \\
  Ours w/o TODO forest & 51 $\pm$ 5 & 94.7 $\pm$ 3.1 \\
  Ours w/o Planner & 40 $\pm$ 3 & 89.3 $\pm$ 2.3 \\
  Ours w/o Extractor & 81 $\pm$ 4 & 81.8 $\pm$ 1.5 \\
  
  \hline
  ReAct & 37 $\pm$ 5 & 77.5 $\pm$ 0.8 \\
  \hline
  \end{tabular}
  \vspace{-0mm}
\end{table*}

We evaluate efficiency from two perspectives: the cost of constructing the knowledge graph and the reduction in execution steps for downstream agents.
First, regarding graph construction, Figure~\ref{fig:coverage} shows that \autocontext{} rapidly covers over $95\%$ of environmental entities—requiring only $200$ steps in TextWorld and $80$ steps in ALFWorld.
In contrast, baselines like IGE require significantly larger step budgets to reach stable performance (e.g., $>160$ steps in ALFWorld), and achieve low success rates at the budget where \autocontext{} has already completed exploration.

Second, regarding execution efficiency, Table~\ref{tab:avg steps} reports the average steps required for successful runs.
Equipped with the instance context, all baselines show a marked reduction in step requirements.
These results demonstrate that the one-off cost of graph construction is effectively amortized, allowing agents to avoid redundant exploration and focus on task execution with improved efficiency.

\paragraph{RQ3: What is the contribution of each component of \autocontext{}?}

To isolate the impact of individual components within \autocontext{}, we conduct ablation studies by systematically removing or simplifying them and evaluating the resulting performance of ReAct. The results, summarized in Table~\ref{tab:ablation}, compare the full model against three variants: (1) \emph{w/o TODO Forest}, where the Planner generates TODO paths based only on the current knowledge graph and recent trajectory, without access to prior attempts; (2) \emph{w/o Planner}, where the Planner is replaced with random actions, removing targeted exploration; and (3) \emph{w/o Extractor}, where the Extractor is replaced with a naive LLM prompt that rewrites the free-form instance-level context instead of maintaining a structured knowledge graph.

The ablations expose distinct failure modes. (1) \textit{w/o TODO Forest}: Removing the TODO forest causes pronounced degradation in \textsc{TextWorld}. Without a mechanism to track visited states and resume exploration, the agent struggles to navigate the long-horizon, maze-like structure, which severely prevents effective environment coverage. (2) \textit{w/o Planner}: Eliminating the Planner results in a performance drop. Without heuristic guidance, unguided exploration accumulates information inefficiently, yielding a context that is insufficient for solving tasks. (3) \textit{w/o Extractor}: Replacing the Extractor leads to performance degradation in \textsc{ALFWorld}. Since this environment generates observations filled with irrelevant details, the unstructured variant records raw, verbose text. This obscures the valid information required for the task.

\section{Conclusion}
\label{sec:conclusion}

This paper presents \autocontext{}, a novel method for instance-level context learning that transforms systematic exploration into persistent instance knowledge. Our method serves as a plug-and-play component for downstream agents. The generated context provides a holistic view that supports planning and decision-making. By constructing this rich and portable instance context, our approach decouples exploration from task execution, and substantially improves downstream agents' success rates and execution efficiency. Extensive experiments across diverse environments demonstrate that these gains hold consistently for various downstream agents.

\section{Limitations}
\label{sec:limitations}

Our approach relies on a one-off exploration to build an instance context, which assumes a relatively stable environment. In dynamic settings where the environment changes frequently, the pre-computed context may become outdated. Since the knowledge graph naturally supports updates, we plan to explore online adjustment mechanisms in future work. Additionally, the exploration process introduces an upfront cost, making our method less suitable for disposable environments that are discarded after a single use.

\bibliographystyle{iclr2026_conference}
\bibliography{reference}

@inproceedings{yao2023react,
  title     = {{ReAct}: Synergizing Reasoning and Acting in Language Models},
  author    = {Yao, Shunyu and Zhao, Jeffrey and Yu, Dian and Du, Nan and Shafran, Izhak and Narasimhan, Karthik and Cao, Yuan},
  booktitle = {International Conference on Learning Representations (ICLR) },
  year      = {2023},
  html      = {https://arxiv.org/abs/2210.03629}
}

@inproceedings{chenscaling,
  title     = {Scaling Autonomous Agents via Automatic Reward Modeling And Planning},
  author    = {Chen, Zhenfang and Chen, Delin and Sun, Rui and Liu, Wenjun and Gan, Chuang},
  booktitle = {The Thirteenth International Conference on Learning Representations},
  year      = {2025}
}

@article{chen2024automanual,
  title   = {Automanual: Constructing instruction manuals by llm agents via interactive environmental learning},
  author  = {Chen, Minghao and Li, Yihang and Yang, Yanting and Yu, Shiyu and Lin, Binbin and He, Xiaofei},
  journal = {Advances in Neural Information Processing Systems},
  volume  = {37},
  pages   = {589--631},
  year    = {2024}
}

@inproceedings{zhao2024expel,
  title     = {Expel: Llm agents are experiential learners},
  author    = {Zhao, Andrew and Huang, Daniel and Xu, Quentin and Lin, Matthieu and Liu, Yong-Jin and Huang, Gao},
  booktitle = {Proceedings of the AAAI Conference on Artificial Intelligence},
  volume    = {38},
  number    = {17},
  pages     = {19632--19642},
  year      = {2024}
}

@article{fu2024autoguide,
  title   = {Autoguide: Automated generation and selection of context-aware guidelines for large language model agents},
  author  = {Fu, Yao and Kim, Dong-Ki and Kim, Jaekyeom and Sohn, Sungryull and Logeswaran, Lajanugen and Bae, Kyunghoon and Lee, Honglak},
  journal = {Advances in Neural Information Processing Systems},
  volume  = {37},
  pages   = {119919--119948},
  year    = {2024}
}

@article{wangvoyager,
  title   = {Voyager: An Open-Ended Embodied Agent with Large Language Models},
  author  = {Wang, Guanzhi and Xie, Yuqi and Jiang, Yunfan and Mandlekar, Ajay and Xiao, Chaowei and Zhu, Yuke and Fan, Linxi and Anandkumar, Anima},
  journal = {Transactions on Machine Learning Research},
  volume  = {2024},
  year    = {2024}
}

@inproceedings{luintelligent,
  title     = {Intelligent Go-Explore: Standing on the Shoulders of Giant Foundation Models},
  author    = {Lu, Cong and Hu, Shengran and Clune, Jeff},
  booktitle = {The Thirteenth International Conference on Learning Representations},
  year      = {2025}
}

@inproceedings{AgentRM,
  author    = {Yu Xia and
               Jingru Fan and
               Weize Chen and
               Siyu Yan and
               Xin Cong and
               Zhong Zhang and
               Yaxi Lu and
               Yankai Lin and
               Zhiyuan Liu and
               Maosong Sun},
  title     = {AgentRM: Enhancing Agent Generalization with Reward Modeling},
  booktitle = {Proceedings of the 63rd Annual Meeting of the Association for Computational
               Linguistics,
               July 27 - August 1, 2025},
  pages     = {19277--19290},
  year      = {2025}
}

@inproceedings{gao2025efficient,
  title     = {An Efficient Context-Dependent Memory Framework for LLM-Centric Agents},
  author    = {Gao, Pengyu and Zhao, Jinming and Chen, Xinyue and Yilin, Long},
  booktitle = {Proceedings of the 2025 Conference of the Nations of the Americas Chapter of the Association for Computational Linguistics: Human Language Technologies (Volume 3: Industry Track)},
  pages     = {1055--1069},
  year      = {2025}
}

@inproceedings{zhu2024bootstrapping,
  title     = {Bootstrapping cognitive agents with a large language model},
  author    = {Zhu, Feiyu and Simmons, Reid},
  booktitle = {Proceedings of the AAAI Conference on Artificial Intelligence},
  volume    = {38},
  number    = {1},
  pages     = {655--663},
  year      = {2024}
}

@inproceedings{synworld,
  title     = {{S}yn{W}orld: Virtual Scenario Synthesis for Agentic Action Knowledge Refinement},
  author    = {Fang, Runnan  and
               Wang, Xiaobin  and
               Liang, Yuan  and
               Qiao, Shuofei  and
               Wu, Jialong  and
               Xi, Zekun  and
               Zhang, Ningyu  and
               Jiang, Yong  and
               Xie, Pengjun  and
               Huang, Fei  and
               Chen, Huajun},
  booktitle = {Proceedings of the 63rd Annual Meeting of the Association for Computational Linguistics (Volume 2: Short Papers)},
  year      = {2025},
  pages     = {437--448}
}

@article{zhu2023ghost,
  title   = {Ghost in the minecraft: Generally capable agents for open-world environments via large language models with text-based knowledge and memory},
  author  = {Zhu, Xizhou and Chen, Yuntao and Tian, Hao and Tao, Chenxin and Su, Weijie and Yang, Chenyu and Huang, Gao and Li, Bin and Lu, Lewei and Wang, Xiaogang and others},
  journal = {arXiv preprint arXiv:2305.17144},
  year    = {2023}
}

@article{wang2024jarvis,
  title     = {Jarvis-1: Open-world multi-task agents with memory-augmented multimodal language models},
  author    = {Wang, Zihao and Cai, Shaofei and Liu, Anji and Jin, Yonggang and Hou, Jinbing and Zhang, Bowei and Lin, Haowei and He, Zhaofeng and Zheng, Zilong and Yang, Yaodong and others},
  journal   = {IEEE Transactions on Pattern Analysis and Machine Intelligence},
  year      = {2024},
  publisher = {IEEE}
}

@inproceedings{zhangomni,
  title     = {OMNI: Open-endedness via Models of human Notions of Interestingness},
  author    = {Zhang, Jenny and Lehman, Joel and Stanley, Kenneth and Clune, Jeff},
  booktitle = {The Twelfth International Conference on Learning Representations}
}

@inproceedings{shridharalfworld,
  author    = {Mohit Shridhar and
               Xingdi Yuan and
               Marc{-}Alexandre C{\^{o}}t{\'{e}} and
               Yonatan Bisk and
               Adam Trischler and
               Matthew J. Hausknecht},
  title     = {ALFWorld: Aligning Text and Embodied Environments for Interactive
               Learning},
  booktitle = {International Conference on Learning Representations},
  year      = {2021}
}

@inproceedings{hafnerbenchmarking,
  author    = {Danijar Hafner},
  title     = {Benchmarking the Spectrum of Agent Capabilities},
  booktitle = {International Conference on Learning Representations},
  year      = {2022}
}

@inproceedings{cote2018textworld,
  title        = {Textworld: A learning environment for text-based games},
  author       = {C{\^o}t{\'e}, Marc-Alexandre and K{\'a}d{\'a}r, Akos and Yuan, Xingdi and Kybartas, Ben and Barnes, Tavian and Fine, Emery and Moore, James and Hausknecht, Matthew and El Asri, Layla and Adada, Mahmoud and others},
  booktitle    = {Workshop on Computer Games},
  pages        = {41--75},
  year         = {2018},
  organization = {Springer}
}

@inproceedings{du2023guiding,
  title        = {Guiding pretraining in reinforcement learning with large language models},
  author       = {Du, Yuqing and Watkins, Olivia and Wang, Zihan and Colas, C{\'e}dric and Darrell, Trevor and Abbeel, Pieter and Gupta, Abhishek and Andreas, Jacob},
  booktitle    = {International Conference on Machine Learning},
  pages        = {8657--8677},
  year         = {2023},
  organization = {PMLR}
}

@inproceedings{hafnermastering,
  title     = {Mastering Atari with Discrete World Models},
  author    = {Hafner, Danijar and Lillicrap, Timothy P and Norouzi, Mohammad and Ba, Jimmy},
  booktitle = {International Conference on Learning Representations},
  year         = {2021},
}

@article{qiao2024agent,
  title   = {Agent planning with world knowledge model},
  author  = {Qiao, Shuofei and Fang, Runnan and Zhang, Ningyu and Zhu, Yuqi and Chen, Xiang and Deng, Shumin and Jiang, Yong and Xie, Pengjun and Huang, Fei and Chen, Huajun},
  journal = {Advances in Neural Information Processing Systems},
  volume  = {37},
  pages   = {114843--114871},
  year    = {2024}
}

@article{shinn2023reflexion,
  title   = {Reflexion: Language agents with verbal reinforcement learning},
  author  = {Shinn, Noah and Cassano, Federico and Gopinath, Ashwin and Narasimhan, Karthik and Yao, Shunyu},
  journal = {Advances in Neural Information Processing Systems},
  volume  = {36},
  pages   = {8634--8652},
  year    = {2023}
}

@misc{openai2025gpt41,
  author       = {OpenAI},
  title        = {Introducing GPT-4.1 in the API},
  howpublished = {\url{https://openai.com/index/gpt-4-1/}},
  year         = {2025},
  month        = apr
}

@article{liu2024deepseek,
  title   = {Deepseek-v3 technical report},
  author  = {Liu, Aixin and Feng, Bei and Xue, Bing and Wang, Bingxuan and Wu, Bochao and Lu, Chengda and Zhao, Chenggang and Deng, Chengqi and Zhang, Chenyu and Ruan, Chong and others},
  journal = {arXiv preprint arXiv:2412.19437},
  year    = {2024}
}

@inproceedings{chae2025web,
  title     = {Web Agents with World Models: Learning and Leveraging Environment Dynamics in Web Navigation},
  author    = {Hyungjoo Chae and Namyoung Kim and Kai Tzu-iunn Ong and Minju Gwak and Gwanwoo Song and Jihoon Kim and Sunghwan Kim and Dongha Lee and Jinyoung Yeo},
  booktitle = {The Thirteenth International Conference on Learning Representations},
  year      = {2025},
  url       = {https://openreview.net/forum?id=moWiYJuSGF}
}

@inproceedings{hao-etal-2023-reasoning,
  title     = {Reasoning with Language Model is Planning with World Model},
  author    = {Hao, Shibo  and
               Gu, Yi  and
               Ma, Haodi  and
               Hong, Joshua  and
               Wang, Zhen  and
               Wang, Daisy  and
               Hu, Zhiting},
  editor    = {Bouamor, Houda  and
               Pino, Juan  and
               Bali, Kalika},
  booktitle = {Proceedings of the 2023 Conference on Empirical Methods in Natural Language Processing},
  month     = dec,
  year      = {2023},
  address   = {Singapore},
  publisher = {Association for Computational Linguistics},
  url       = {https://aclanthology.org/2023.emnlp-main.507/},
  doi       = {10.18653/v1/2023.emnlp-main.507},
  pages     = {8154--8173}
}

@inproceedings{kirk2024improving,
  title     = {Improving knowledge extraction from llms for task learning through agent analysis},
  author    = {Kirk, James R and Wray, Robert E and Lindes, Peter and Laird, John E},
  booktitle = {Proceedings of the AAAI Conference on Artificial Intelligence},
  volume    = {38},
  number    = {16},
  pages     = {18390--18398},
  year      = {2024}
}

@article{huang2024wese,
  title   = {Wese: Weak exploration to strong exploitation for llm agents},
  author  = {Huang, Xu and Liu, Weiwen and Chen, Xiaolong and Wang, Xingmei and Lian, Defu and Wang, Yasheng and Tang, Ruiming and Chen, Enhong},
  journal = {arXiv preprint arXiv:2404.07456},
  year    = {2024}
}

@inproceedings{xiong2024watch,
  title     = {Watch Every Step! LLM Agent Learning via Iterative Step-level Process Refinement},
  author    = {Xiong, Weimin and Song, Yifan and Zhao, Xiutian and Wu, Wenhao and Wang, Xun and Wang, Ke and Li, Cheng and Peng, Wei and Li, Sujian},
  booktitle = {Proceedings of the 2024 Conference on Empirical Methods in Natural Language Processing},
  pages     = {1556--1572},
  year      = {2024}
}

@article{guan2024amor,
  title   = {Amor: A recipe for building adaptable modular knowledge agents through process feedback},
  author  = {Guan, Jian and Wu, Wei and Xu, Peng and Wang, Hongning and Huang, Minlie and others},
  journal = {Advances in Neural Information Processing Systems},
  volume  = {37},
  pages   = {126118--126148},
  year    = {2024}
}

@inproceedings{holt2025improving,
  title     = {Improving LLM Agent Planning with In-Context Learning via Atomic Fact Augmentation and Lookahead Search},
  author    = {Holt, Samuel and Luyten, Max Ruiz and Pouplin, Thomas and van der Schaar, Mihaela},
  year={2025},
  booktitle = {ICML 2025 Workshop on Computer Use Agents}
}

@article{lin2023swiftsage,
  title   = {Swiftsage: A generative agent with fast and slow thinking for complex interactive tasks},
  author  = {Lin, Bill Yuchen and Fu, Yicheng and Yang, Karina and Brahman, Faeze and Huang, Shiyu and Bhagavatula, Chandra and Ammanabrolu, Prithviraj and Choi, Yejin and Ren, Xiang},
  journal = {Advances in Neural Information Processing Systems},
  volume  = {36},
  pages   = {23813--23825},
  year    = {2023}
}

@inproceedings{prasad2024adapt,
  title     = {ADaPT: As-Needed Decomposition and Planning with Language Models},
  author    = {Prasad, Archiki and Koller, Alexander and Hartmann, Mareike and Clark, Peter and Sabharwal, Ashish and Bansal, Mohit and Khot, Tushar},
  booktitle = {Findings of the Association for Computational Linguistics: NAACL 2024},
  pages     = {4226--4252},
  year      = {2024}
}

@inproceedings{wang2024soft,
  title     = {Soft Self-Consistency Improves Language Models Agents},
  author    = {Wang, Han and Prasad, Archiki and Stengel-Eskin, Elias and Bansal, Mohit},
  booktitle = {Proceedings of the 62nd Annual Meeting of the Association for Computational Linguistics (Volume 2: Short Papers)},
  pages     = {287--301},
  year      = {2024}
}

@inproceedings{zhaoempowering,
  title     = {Empowering Large Language Model Agents through Action Learning},
  author    = {Zhao, Haiteng and Ma, Chang and Wang, Guoyin and Su, Jing and Kong, Lingpeng and Xu, Jingjing and Deng, Zhi-Hong and Yang, Hongxia},
  booktitle = {First Conference on Language Modeling},
  year      = {2024}
}

@inproceedings{zhu2025knowagent,
  title     = {KnowAgent: Knowledge-Augmented Planning for LLM-Based Agents},
  author    = {Zhu, Yuqi and Qiao, Shuofei and Ou, Yixin and Deng, Shumin and Lyu, Shiwei and Shen, Yue and Liang, Lei and Gu, Jinjie and Chen, Huajun and Zhang, Ningyu},
  booktitle = {Findings of the Association for Computational Linguistics: NAACL 2025},
  pages     = {3709--3732},
  year      = {2025}
}

@inproceedings{song2024trial,
  title     = {Trial and Error: Exploration-Based Trajectory Optimization of LLM Agents},
  author    = {Song, Yifan and Yin, Da and Yue, Xiang and Huang, Jie and Li, Sujian and Lin, Bill Yuchen},
  booktitle = {Proceedings of the 62nd Annual Meeting of the Association for Computational Linguistics (Volume 1: Long Papers)},
  pages     = {7584--7600},
  year      = {2024}
}

@article{wu2023spring,
  title   = {Spring: Studying papers and reasoning to play games},
  author  = {Wu, Yue and Min, So Yeon and Prabhumoye, Shrimai and Bisk, Yonatan and Salakhutdinov, Russ R and Azaria, Amos and Mitchell, Tom M and Li, Yuanzhi},
  journal = {Advances in Neural Information Processing Systems},
  volume  = {36},
  pages   = {22383--22687},
  year    = {2023}
}

@inproceedings{brohan2023can,
  title        = {Do as i can, not as i say: Grounding language in robotic affordances},
  author       = {Brohan, Anthony and Chebotar, Yevgen and Finn, Chelsea and Hausman, Karol and Herzog, Alexander and Ho, Daniel and Ibarz, Julian and Irpan, Alex and Jang, Eric and Julian, Ryan and others},
  booktitle    = {Conference on robot learning},
  pages        = {287--318},
  year         = {2023},
  organization = {PMLR}
}

@article{schick2023toolformer,
  title   = {Toolformer: Language models can teach themselves to use tools},
  author  = {Schick, Timo and Dwivedi-Yu, Jane and Dess{\`\i}, Roberto and Raileanu, Roberta and Lomeli, Maria and Hambro, Eric and Zettlemoyer, Luke and Cancedda, Nicola and Scialom, Thomas},
  journal = {Advances in Neural Information Processing Systems},
  volume  = {36},
  pages   = {68539--68551},
  year    = {2023}
}

@inproceedings{ammanabrolugraph,
  title     = {Graph Constrained Reinforcement Learning for Natural Language Action Spaces},
  author    = {Ammanabrolu, Prithviraj and Hausknecht, Matthew},
  booktitle = {International Conference on Learning Representations},
  year      = {2020}
}

@inproceedings{golchha2024language,
  title     = {Language Guided Exploration for RL Agents in Text Environments},
  author    = {Golchha, Hitesh and Yerawar, Sahil and Patel, Dhruvesh and Dan, Soham and Murugesan, Keerthiram},
  booktitle = {Findings of the Association for Computational Linguistics: NAACL 2024},
  pages     = {93--102},
  year      = {2024}
}

@inproceedings{basavatia2024starling,
  title     = {STARLING: Self-supervised Training of Text-based Reinforcement Learning Agent with Large Language Models},
  author    = {Basavatia, Shreyas and Murugesan, Keerthiram and Ratnakar, Shivam},
  booktitle = {Findings of the Association for Computational Linguistics ACL 2024},
  pages     = {15804--15819},
  year      = {2024}
}

@article{yang2024react,
  title   = {React meets actre: When language agents enjoy training data autonomy},
  author  = {Yang, Zonghan and Li, Peng and Yan, Ming and Zhang, Ji and Huang, Fei and Liu, Yang},
  journal = {arXiv preprint arXiv:2403.14589},
  year    = {2024}
}

@article{yao2023tree,
  title   = {Tree of thoughts: Deliberate problem solving with large language models},
  author  = {Yao, Shunyu and Yu, Dian and Zhao, Jeffrey and Shafran, Izhak and Griffiths, Tom and Cao, Yuan and Narasimhan, Karthik},
  journal = {Advances in neural information processing systems},
  volume  = {36},
  pages   = {11809--11822},
  year    = {2023}
}

@inproceedings{kagaya2024rap,
  title     = {RAP: Retrieval-Augmented Planning with Contextual Memory for Multimodal LLM Agents},
  author    = {Kagaya, Tomoyuki and Yuan, Thong Jing and Lou, Yuxuan and Karlekar, Jayashree and Pranata, Sugiri and Kinose, Akira and Oguri, Koki and Wick, Felix and You, Yang},
  booktitle = {NeurIPS 2024 Workshop on Open-World Agents},
  year      = 2024
}

@inproceedings{AnokhinSSEK0B25,
  author    = {Petr Anokhin and
               Nikita Semenov and
               Artyom Y. Sorokin and
               Dmitry Evseev and
               Andrey Kravchenko and
               Mikhail Burtsev and
               Evgeny Burnaev},
  title     = {AriGraph: Learning Knowledge Graph World Models with Episodic Memory
               for {LLM} Agents},
  booktitle = {Proceedings of the Thirty-Fourth International Joint Conference on
               Artificial Intelligence, {IJCAI} 2025, Montreal, Canada, August 16-22,
               2025},
  pages     = {12--20},
  year      = {2025}
}

@inproceedings{hu2025hiagent,
  title={Hiagent: Hierarchical working memory management for solving long-horizon agent tasks with large language model},
  author={Hu, Mengkang and Chen, Tianxing and Chen, Qiguang and Mu, Yao and Shao, Wenqi and Luo, Ping},
  booktitle={Proceedings of the 63rd Annual Meeting of the Association for Computational Linguistics (Volume 1: Long Papers)},
  pages={32779--32798},
  year={2025}
}

@article{yao2022webshop,
  title={Webshop: Towards scalable real-world web interaction with grounded language agents},
  author={Yao, Shunyu and Chen, Howard and Yang, John and Narasimhan, Karthik},
  journal={Advances in Neural Information Processing Systems},
  volume={35},
  pages={20744--20757},
  year={2022}
}

@article{xie2024osworld,
  title={Osworld: Benchmarking multimodal agents for open-ended tasks in real computer environments},
  author={Xie, Tianbao and Zhang, Danyang and Chen, Jixuan and Li, Xiaochuan and Zhao, Siheng and Cao, Ruisheng and Hua, Toh J and Cheng, Zhoujun and Shin, Dongchan and Lei, Fangyu and others},
  journal={Advances in Neural Information Processing Systems},
  volume={37},
  pages={52040--52094},
  year={2024}
}

@article{guo2025seed1,
  title={Seed1. 5-vl technical report},
  author={Guo, Dong and Wu, Faming and Zhu, Feida and Leng, Fuxing and Shi, Guang and Chen, Haobin and Fan, Haoqi and Wang, Jian and Jiang, Jianyu and Wang, Jiawei and others},
  journal={arXiv preprint arXiv:2505.07062},
  year={2025}
}

@article{yang2023intercode,
  title={Intercode: Standardizing and benchmarking interactive coding with execution feedback},
  author={Yang, John and Prabhakar, Akshara and Narasimhan, Karthik and Yao, Shunyu},
  journal={Advances in Neural Information Processing Systems},
  volume={36},
  pages={23826--23854},
  year={2023}
}

@inproceedings{chaeweb,
  title={Web Agents with World Models: Learning and Leveraging Environment Dynamics in Web Navigation},
  author={Chae, Hyungjoo and Kim, Namyoung and Ong, Kai Tzu-iunn and Gwak, Minju and Song, Gwanwoo and Kim, Jihoon and Kim, Sunghwan and Lee, Dongha and Yeo, Jinyoung},
  booktitle={The Thirteenth International Conference on Learning Representations},
  year=2025
}

\appendix

\renewcommand{\contentsname}{Appendix}

\section*{Appendix Contents}
\startcontents[appendix]
\printcontents[appendix]{}{1}{}

\section{More Related Work}
\label{sec:more-related-work}

\paragraph{World Models for LLM Agents.}
Integrating world models into the reasoning loop of LLM-based agents is an emerging direction. DreamerV2~\citep{hafnermastering} learns a discrete latent dynamics model and achieves human-level Atari performance by planning in its learned state space. \citet{hao-etal-2023-reasoning} argue that effective chain-of-thought reasoning in an LLM agent can be viewed as implicit world-model planning, and propose techniques to align the LLM’s reasoning with a latent world dynamics model. \citet{chaeweb} take a more direct approach by training a separate world-model module that simulates the outcome of the agent’s actions in a web navigation task. These world-model methods emphasize predicting environment dynamics or outcomes, whereas our work focuses on extracting static but critical instance facts.

\paragraph{LLM Agent Frameworks.}
Various agent frameworks and techniques introduce architectural improvements to better coordinate an LLM’s reasoning and acting~\citep{yao2023react, yao2023tree, lin2023swiftsage, prasad2024adapt, yang2024react, xiong2024watch, zhu2024bootstrapping, wang2024soft, schick2023toolformer, brohan2023can, zhaoempowering}. ReAct~\citep{yao2023react} uses chain-of-thought reasoning steps with actions to enable more coherent and informed decisions. SwiftSage~\citep{lin2023swiftsage} expands on this idea by combining fast reactive thinking for straightforward steps with slow deliberative planning for more complex decisions. ADaPT~\citep{prasad2024adapt} proposes an on-demand task decomposition planner. It attempts high-level plans but if the agent gets stuck on a subtask, the method recursively breaks that subtask down further, dynamically adjusting the plan hierarchy to the LLM’s capabilities and the task complexity. Our AutoContext approach is orthogonal to these agent architectures. Rather than altering how an agent plans or executes, we provide a plug-in knowledge document that any of these agents can leverage to boost their performance in a new instance.

\paragraph{Open-ended World Agent.}
There exists a line of work~\citep{wangvoyager,zhu2023ghost,wang2024jarvis} on developing capable agents in open-ended environments such as Minecraft. Ghost in the Minecraft (GITM)~\citep{zhu2023ghost} introduces a hierarchical framework that integrates large language models with text-based knowledge and memory to decompose long-horizon goals into structured actions. JARVIS-1~\citep{wang2024jarvis} couples a multimodal language model with a memory mechanism, enabling self-improvement through lifelong learning.

\section{Prompt Templates}
\label{sec:prompt}

\begin{logbox}{Planner: Propose TODOs}
## Objective

You propose action sequences to gather information from an interactive environment.
- Correct wrong actions: You scan existing action sequences for failed actions and avoid repeating them.
- Dependency resolution: You analyze the knowledge document and the action sequences to propose plausible action sequences to explore the environment.
- Explore efficiently: Analyze the existing action sequences for successful attempts. Check the current knowledge document for knowledge gaps. Try to fill these gaps by mimicking the successful attempts.

- Your action sequences should follow this format: `state_name -> action -> action -> ...`
- The state must be any of the states: {{ state_list }}.

## Background

{{ background_for_the_environment }}

## Existing Action Sequences

{{ todo_forest }}

## Knowledge Document

{{ knowledge }}

## Output Format

First, analyze step by step.

Then provide your action sequences by strictly following the format below.

<thought>
Provide your step-by-step analysis here.
</thought>
<action_sequences>
action sequence 1
action sequence 2
...
</action_sequences>
\end{logbox}

\begin{logbox}{Planner: Promote States}
## Objective

An agent is exploring an interactive environment. You will be given a list of its trajectories. Each trajectory starts with a state.

Your objective is to choose one trajectory to create a new state. The new state must be significantly different from existing states.

## Background

{{ background_for_the_environment }}

## Existing States

{{ state_list }}

## Trajectories

{{ todo_forest }}

## Output Format

First, analyze step by step.

Then, provide your response by strictly following the format below.

```json
{
  "trajectory": idx of the selected trajectory,
  "state_name": "descriptive name for the new state",
  "summary": "summary of the facts about this state"
}
```
\end{logbox}

\begin{logbox}{Actor: Subtask}
## Objective

You control an agent in an interactive environment. The agent can perform various actions in the environment.

Your aim is to complete the task: {{ task }}.

## Guidelines

- Strategic Planning: Plan your actions strategically to efficiently explore and complete the task, but remain flexible to pivot when new information emerges.
- Adaptive Learning: Pay attention to your recent action results and adapt your strategy accordingly.
- After completing the task, output `Complete Task` as your action to finish the task.
- Do not give up. Try different ways to complete your task. Some actions have preconditions. Analyze the current knowledge document and find possible solutions to complete your task.

## Background

{{ background_for_the_environment }}

## Knowledge Document

Below is some knowledge about the environment. You may use it to help you complete the task. However, remember that the knowledge can be incorrect or incomplete. You still need to find the correct way to complete your task.

{{ knowledge }}

## Output Format

Provide your response by strictly following the format below.

<thought>
Analyze briefly step by step here.
</thought>
<action>
Your next action here
</action>
\end{logbox}

\begin{logbox}{Extractor}
## Objective

You are an expert in analyzing an LLM agent's trajectory.

An agent is operating in an interactive environment. You will be given:
- the trajectory of the agent.
- the current knowledge graph about the environment.

Your task is to analyze the trajectory step by step and propose modifications to the knowledge graph:
- add new triples that are supported by the trajectory.
- delete triples that are contradicted by the trajectory.
If no modification is needed, return empty lists for both `add` and `delete`.

## Background

{{ background_for_the_environment }}

## Current Knowledge Graph

The knowledge is a list of lines; each line is a triple: subject, relation, object.

{{ knowledge }}

## Trajectory

{{ trajectory }}

## Output Format

Output a JSON object inside a `json` code block, with the following structure:

```json
{
  "thought": "your brief analysis here",
  "add": [
    ["subject", "relation", "object"]
  ],
  "delete": [
    ["subject", "relation", "object"]
  ]
}
```
\end{logbox}

\section{Case Study}
\label{sec:case study}

We present examples from the \textsc{Crafter} experiments generated by \autocontext{}.

\subsection{Example of TODO Forest}

This TODO forest illustrates how \autocontext{} handles trial and error. In the representation below, nodes and key results are separated by a colon. For clarity, each failure and correction pair is marked with a different color.

\autocontext{} initially encounters failures when placing a furnace and making a sword due to insufficient resources caused by prior consumption. After failure, \autocontext{} attempts alternative ways by generating corrected paths that include necessary resource gathering (i.e., stone and wood). This allows the agent to satisfy the preconditions and successfully complete the tasks. Parts of the TODO forest and branches of the trees are omitted for brevity.

\begin{logbox}{Example of TODO Forest in Crafter}
...

- init_state: Agent's position: [32, 32]. Nearby objects: tree.
  - chop tree: succeed. Nearby objects: coal, iron, path, stone, tree.
    - make wood pickaxe: succeed. Nearby objects: coal, iron, path, stone, table, tree. [reach state: crafted_wood_pickaxe]
      - move to stone area: succeed. Nearby objects: path, stone, tree.
        - mine stone: succeed. Nearby objects: path, stone, tree.

- crafted_wood_pickaxe: Agent has crafted a wood pickaxe and a table has appeared. Nearby objects: coal, iron, path, stone, table, tree.
  - mine stone: succeed. Nearby objects: coal, iron, path, stone, table, tree.
    - make stone pickaxe: succeed. Nearby objects: coal, iron, path, stone, table, tree. [reach state: crafted_stone_pickaxe]
      (*@\hlgreen{- place furnace: fail. Nearby objects: path, stone, table, tree.}@*)

- crafted_stone_pickaxe: Agent has crafted a stone pickaxe. Nearby objects: coal, iron, path, stone, table, tree.
  - collect iron: succeed. Nearby objects: coal, path, stone, table, tree.
    - collect wood: succeed. Nearby objects: coal, path, sand, stone, table, tree, water.
      (*@\hlblue{- make stone sword: fail. Nearby objects: path, stone, tree.}@*)

- crafted_wood_sword: Agent has crafted a wood sword. Nearby objects: coal, iron, path, sand, stone, table, tree, water.
  - collect stone: succeed. Nearby objects: coal, iron, path, sand, stone, table, water.
    - collect coal: succeed. Nearby objects: coal, iron, path, stone, table, water.
      (*@\hlgreen{- place furnace: succeed. Nearby objects: coal, furnace, path, stone, table. [reach state: placed\_furnace]}@*)
    - collect wood: succeed. Nearby objects: path, stone, tree.
      (*@\hlblue{- make stone sword: succeed. Nearby objects: coal, iron, path, sand, stone, table, water. [reach state: crafted\_stone\_sword]}@*)
      
...
\end{logbox}

\subsection{Example of Instance Context}

Below is an instance context generated by \autocontext{}, represented as a collection of knowledge graph triplets.

\begin{logbox}{}
Make Stone Pickaxe, requires, 1 stone
Make Stone Pickaxe, requires, table
Make Stone Pickaxe, unlocks, achievement make_stone_pickaxe
Make Stone Sword, requires, 1 stone
Make Stone Sword, requires, 1 wood
Make Stone Sword, requires, table
Make Stone Sword, unlocks, achievement make_stone_sword
Make Wood Pickaxe, requires, 1 wood
Make Wood Pickaxe, requires, table
Make Wood Pickaxe, unlocks, achievement make_wood_pickaxe
Make Wood Sword, requires, 1 wood
Make Wood Sword, requires, table
Make Wood Sword, unlocks, achievement make_wood_sword
Place Furnace, requires, 4 stone
Place Furnace, requires, passable front tile
Place Furnace, unlocks, achievement place_furnace
Place Stone, requires, 1 stone
Place Stone, unlocks, achievement place_stone
Place Table, requires, 2 wood
Place Table, requires, passable front tile
Place Table, unlocks, achievement place_table
do to coal, requires, wood_pickaxe
do to coal, unlocks, achievement collect_coal
do to coal, yields, 1 coal
do to cow, unlocks, achievement eat_cow
do to iron, unlocks, achievement collect_iron
do to iron, yields, 1 iron
do to skeleton, unlocks, achievement defeat_skeleton
do to stone, requires, wood_pickaxe
do to stone, unlocks, achievement collect_stone
do to stone, yields, 1 stone
do to tree, unlocks, achievement collect_wood
do to tree, yields, 1 wood
do to water, unlocks, achievement collect_drink
do to water, yields, drink
location [26,31], has around, grass, tree, stone, coal, cow, path
location [28,26], has around, coal, grass, path, sand, stone, tree, water
location [33,20], has around, grass, tree, sand, cow
location [35,27], has around, grass, tree
location [40,30], has around, grass, stone, path, sand, tree, coal, water, iron
location [44,28], has around, water, stone, path, sand, lava, coal, iron, skeleton
\end{logbox}

\subsection{Example Trajectory of ReAct + \autocontext{}}

We present a trajectory of a ReAct agent augmented with \autocontext{}. For brevity, we display only the steps that unlock new achievements, omitting intermediate steps. This trajectory demonstrates how the instance context provides a holistic view of the environment, which is important for strategic decision-making. Guided by this global view, the agent effectively balances the difficulty of potential objectives and prioritizes tasks to complete achievements efficiently.

\begin{logbox}{}
[step]: 1
[thought]: I'm currently at position [32, 32] with no items in my inventory. The nearest cow is 4 steps to my south-east, which could help with the "eat_cow" achievement. However, since I have no tools or weapons, I should first focus on gathering basic resources like wood and stone to craft tools. The simplest achievements to start with would be "collect_wood" and "collect_stone", which require a wood pickaxe. But to make a wood pickaxe, I need wood and a table. Since I have no wood yet, my first priority should be to collect wood from trees. Looking at the surroundings, I see grass in all directions. I need to find trees. According to the knowledge, location [26,31] has trees, so I should head in that direction (west and slightly north). I'll start by moving west to get closer to that area.
[action]: Move West

...

[step]: 6
[thought]: I've successfully reached a tree at [28,31] and am facing it. Now is the perfect time to collect wood. Using the "Do" action on the tree will give me wood and complete the "collect_wood" achievement. This is my first resource collection, which is crucial for progressing toward crafting tools. After collecting wood, my next steps should be: 1. Collect more wood (I'll need at least 2 wood to make a table) 2. Find stone to make tools (wood pickaxe requires 1 wood + table) 3. Alternatively, I could try to make a wood pickaxe first if I find stone nearby But first, let's collect this wood.
[action]: Do
[observation]: You collected wood. You unlocked new achievement: collect_wood!

...

[step]: 10
[thought]: Excellent! I now have 2 wood in my inventory. This means I can now place a table, which is a crucial step toward crafting tools. Next steps: 1. Place a table (requires 2 wood) - this will consume my wood but unlock "place_table" achievement 2. After placing the table, I can craft a wood pickaxe (requires 1 wood) 3. But I'll need to collect more wood first (since placing the table will use up my current 2 wood) However, looking at my surroundings: - There's another tree 2 steps north - Stone is 3 steps north-west (will need pickaxe to collect) Best immediate action: 1. First, place the table to unlock that achievement 2. Then collect more wood from the nearby tree to craft tools 3. I should place the table in a clear spot - current position [26,31] seems suitable Let's place the table now.
[action]: Place Table
[observation]: You placed table at your front. You unlocked new achievement: place_table!

...

[step]: 16
[thought]: It seems I'm blocked from moving south directly to the table. However, since I'm facing the table (south), I can try to interact with it to craft the wood pickaxe without physically moving onto it. The crafting recipe for a wood pickaxe requires: - 1 wood (which I have) - A table (which I'm facing) I'll attempt to craft the wood pickaxe while facing the table. This should: 1. Consume 1 wood 2. Create a wood pickaxe 3. Unlock the "make_wood_pickaxe" achievement 4. Allow me to collect stone afterward
[action]: Make Wood Pickaxe
[observation]: You made wood pickaxe. You unlocked new achievement: make_wood_pickaxe!

...

[step]: 18
[thought]: I'm blocked from moving west to the stone, but since I'm facing it and have a wood pickaxe, I can try to interact with it using the "Do" action to collect the stone. This should work because: 1. I have the required tool (wood pickaxe) 2. I'm facing the stone 3. This will unlock "collect_stone" achievement 4. Stone will be useful for crafting better tools After collecting stone, I can: 1. Craft stone pickaxe (requires 1 stone + table) 2. Or craft stone sword (requires 1 stone + 1 wood + table) 3. But I'll need more wood for either option Let's try collecting the stone now.
[action]: Do
[observation]: You collected stone. You unlocked new achievement: collect_stone!

...

[step]: 38
[thought]: I've moved west to [24,30] and am now facing west with a passable path in front of me at [23,30]. This is a good position to place a stone: 1. I have 2 stones in inventory 2. Facing a passable tile (path) 3. Requirements for "place_stone" are met 4. This will use 1 stone and unlock an achievement Let's place the stone now on the path in front of me.
[action]: Place Stone
[observation]: You placed stone at your front. You unlocked new achievement: place_stone!

...

[step]: 41
[thought]: I've successfully moved north to [24,29] and now have coal directly in front of me at [24,28]. With my wood pickaxe equipped, I can collect this coal to unlock the "collect_coal" achievement. Steps: 1. Confirm I'm facing north (coal is in front) 2. Use "Do" action to collect coal 3. This should give me 1 coal and complete the achievement Let's collect the coal now.
[action]: Do
[observation]: You collected coal. You unlocked new achievement: collect_coal!

...

[step]: 54
[thought]: Perfect! I've reached [24,31] with the table directly in front of me at [25,31]. Now I can craft the wood sword: Requirements: - 1 wood (I have this) - Table (in front of me) - Need to be facing the table With all conditions met, I'll attempt to craft the wood sword now. This should unlock the "make_wood_sword" achievement.
[action]: Make Wood Sword
[observation]: You made wood sword. You unlocked new achievement: make_wood_sword!

...

[step]: 62
[thought]: Great! I'm now at [22,33] with the cow directly in front of me at [21,33]. I'll continue attacking it until it's defeated: 1. Use "Do" action to attack the cow 2. Repeat until cow is defeated 3. Then use "Do" again to eat it 4. This should unlock "eat_cow" achievement Let's attack the cow again.
[action]: Do
[observation]: You hit cow and caused damage. You ate cow. You unlocked new achievement: eat_cow!

...

[step]: 75
[thought]: Perfect! I've reached [23,27] with water directly in front of me at [23,26]. Now I can: 1. Use "Do" action to collect water/drink 2. This should unlock "collect_drink" achievement 3. Replenish my drink level (currently 6/9) Let's collect the water now.
[action]: Do
[observation]: You drank water. You unlocked new achievement: collect_drink!

...

[step]: 86
[thought]: Perfect! I've reached [25,30] with the table directly in front of me at [25,31]. Now I can craft stone tools: 1. First craft stone pickaxe (requires 1 stone) - Will unlock "make_stone_pickaxe" achievement 2. Then craft stone sword (requires 1 stone + 1 wood) - Will unlock "make_stone_sword" achievement 3. I have enough resources for both (4 stone, 1 wood) Let's start by crafting the stone pickaxe.
[action]: Make Stone Pickaxe
[observation]: You made stone pickaxe. You unlocked new achievement: make_stone_pickaxe!

...

[step]: 98
[thought]: I've reached [25,29] with grass in front of me at [25,30]. This is a good spot to place the furnace. To place the furnace: 1. Requires 4 stone (I have exactly 4) 2. Needs a passable front tile (grass at [25,30] is passable) 3. Will unlock "place_furnace" achievement Let's place the furnace now.
[action]: Place Furnace
[observation]: You placed furnace at your front. You unlocked new achievement: place_furnace!

...

[step]: 127
[thought]: I've reached [24,31] with the table directly east at [25,31]. Now I can attempt to craft the stone sword since I have: - 1 wood (required) - 1 stone (required) - Table is available Let's craft the stone sword now.
[action]: Make Stone Sword
[observation]: You made stone sword. You unlocked new achievement: make_stone_sword!

...

[step]: 166
[thought]: I've damaged the zombie but it's still present. I need to continue attacking until it's defeated. Since I have a stone sword, I should: 1. Attack the zombie again with "Do" action 2. Repeat until the zombie is defeated 3. Then proceed south toward the iron location Let's attack the zombie again to finish it off.
[action]: Do
[observation]: You hit zombie and caused damage. You unlocked new achievement: defeat_zombie!

...

[step]: 177
[thought]: I've reached [43,28] with iron directly east at [44,28]. The zombie is still 1 step west at [42,28]. My health is critically low (3/9). Priority actions: 1. Collect iron immediately using "Do" with stone pickaxe 2. This will unlock "collect_iron" achievement 3. Then move north to [43,27] (path) to escape zombie 4. Consider sleeping to restore health if possible Let's collect the iron first while we have the chance.
[action]: Do
[observation]: You collected iron. You unlocked new achievement: collect_iron!

...

[step]: 243
[thought]: My energy has fully restored to 9/9 while sleeping. I'll now attempt to: 1. Use the "Sleep" action again, as full energy might trigger waking 2. If that fails, try moving west onto the path (1 step away) 3. Then immediately address my critical water need (3 steps south) 4. Be prepared to defend against the skeleton (4 steps northeast) with stone_sword This is my final attempt to break the sleep state before terminating.
[action]: Sleep
[observation]: You woke up. You cannot sleep when your energy is full. You unlocked new achievement: wake_up!

...

[step]: 323
[thought]: The "Do" action on grass didn't yield a plant. Next steps: 1. May need to try multiple times to find a plant 2. Will attempt "Do" action 4 more times 3. If still no plant, will move to different grassy area 4. Nearest alternative is [33,20] (also grassy) 5. Keep monitoring zombie (3 steps west) Persisting with attempts to find edible plants here.
[action]: Do
[observation]: You tried `Do` to grass. You collected sapling. You unlocked new achievement: collect_sapling!

...

[step]: 348
[thought]: Now at [39,22] with grass in front. Ready to plant: 1. Will use "Place Plant" action to plant sapling 2. Grass in front is passable - suitable for planting 3. This may unlock "place_plant" achievement 4. Planted sapling may grow into edible plant later 5. Have spare sapling in case first doesn't work Attempting to plant sapling to create food source.
[action]: Place Plant
[observation]: You placed plant at your front. You unlocked new achievement: place_plant!

...

[step]: 394
[thought]: Now in combat position with skeleton directly east. Critical considerations: 1. Skeleton is adjacent - ready to attack 2. Have stone_sword equipped (best available weapon) 3. Extremely dangerous position with lava 1 step south 4. Health is 7/9 - can survive at least one hit 5. Will attack immediately to complete "defeat_skeleton" achievement Initiating combat with skeleton using stone sword.
[action]: Do
[observation]: You hit skeleton and caused damage. You unlocked new achievement: defeat_skeleton!
...
\end{logbox}

\end{document}